%% file: iclr2027_conference.tex
\documentclass{article} 
\input{preamble}

\title{No Channels Left Behind: Understanding Diffusion Acceleration through Residual Gates}

\author{
  Junha Park\textsuperscript{1}\thanks{Done during internship at NAVER AI Lab.}\quad
  Jaemyung Yu\textsuperscript{2}\quad
  Jaehui Hwang\textsuperscript{2}\quad
  Jong-Seok Lee\textsuperscript{1,$\dagger$}\quad
  Dongyoon Han\textsuperscript{2,$\dagger$} \\
  \normalfont \textsuperscript{1}Yonsei University\qquad \textsuperscript{2}NAVER AI Lab \\
  \small{\textsuperscript{$\dagger$}Corresponding authors.}
}

\iclrfinalcopy 
\begin{document}

\maketitle
\lhead{Preprint}

\vspace{-1.5em}
\input{sections/00-abs}
\vspace{-1.5em}
\input{sections/01-intro}
\input{sections/02-prelim}

\input{sections/03-observations}
\input{sections/04-action}
\input{sections/05-discussion}
\input{sections/05-rel}
\input{sections/06-con}

\bibliography{strings_long, iclr2027_conference}
\bibliographystyle{iclr2027_conference}

\appendix
\input{sections/07-reproducibility}
\input{sections/08-radial-analysis}

\end{document}

%% file: preamble.tex
\usepackage{iclr2027_conference,times}

\input{math_commands.tex}

\usepackage{hyperref}
\usepackage{url}

\usepackage{kotex, xspace}
\usepackage[capitalize]{cleveref}
\usepackage[dvipsnames,x11names]{xcolor}
\usepackage{pifont}

\input{figure-colors}
\hypersetup{colorlinks=true,allcolors=primary}
\input{manuscript-figures}
\usepackage[section]{placeins} 
\usepackage{caption}
\usepackage{subcaption}
\usepackage{booktabs}
\usepackage{wrapfig}
\renewcommand{\paragraph}[1]{\vspace{1.25mm}\noindent\textbf{#1}\xspace}

\newcommand{\ourmetric}{\texorpdfstring{$\mathrm{nPR}(\vct{\alpha})$\xspace}{nPR(α)}}
\newcommand{\ourmetricmean}{\texorpdfstring{$\overline{\mathrm{nPR}(\vct{\alpha})}$\xspace}{nPR(α)}}
\newcommand{\ourmethod}{Radial objective\xspace}
\colorlet{TeacherComparison}{black!45}
\newcommand{\teacherentry}[1]{\textcolor{TeacherComparison}{#1}}

\newcommand{\vct}[1]{\boldsymbol{#1}}
\newcommand{\mat}[1]{\boldsymbol{#1}}

\newcommand{\ie}{\textit{i.e.}\xspace}

\usepackage{colortbl}

\usepackage{soul}

\newif\ifshowhl \showhltrue            

\DeclareRobustCommand{\hlc}[2]{%
  \ifshowhl{\sethlcolor{#1}\hl{#2}}\else#2\fi}

\colorlet{myblue}{MidnightBlue!10}
\colorlet{myred}{RubineRed!10}
\colorlet{my2ndblue}{MidnightBlue!5}

\DeclareRobustCommand{\hlr}[1]{\hlc{myred}{#1}}

\DeclareRobustCommand{\hlg}[1]{\hlc{Green!10}{#1}}

\input{radial-analysis-preamble}

%% file: math_commands.tex
\usepackage{amsmath,amsfonts,bm}
\DeclareMathAlphabet{\mathsfit}{OT1}{cmss}{m}{sl}

\newcommand{\figleft}{{\em (Left)}\xspace}
\newcommand{\figcenter}{{\em (Center)}\xspace}
\newcommand{\figright}{{\em (Right)}\xspace}

\newcommand{\captiona}{{\em (a)}\xspace}
\newcommand{\captionb}{{\em (b)}\xspace}
\newcommand{\captionc}{{\em (c)}\xspace}

\def\eqref#1{equation~\ref{#1}}

\def\1{\mathbf{1}}



%% file: figure-colors.tex
\usepackage{xcolor}
\definecolor{Red}{HTML}{E64B35}
\definecolor{Blue}{HTML}{4DBBD5}
\definecolor{Green}{HTML}{00A087}
\definecolor{Navy}{HTML}{3C5488}
\definecolor{Salmon}{HTML}{F39B7F}
\definecolor{Gray}{HTML}{8A8A8A}
\colorlet{primary}{Green}
\colorlet{neutral}{Gray}

%% file: manuscript-figures.tex
\usepackage{graphicx}
\usepackage{tikz} 
\newcommand{\plotfigure}[1]{\includegraphics{figures/pdf/#1.pdf}}

%% file: radial-analysis-preamble.tex
\usepackage{amsthm}
\newcommand{\radialsurrogate}{\widetilde{\mathcal L}_{\mathrm{rad}}}
\newcommand{\radialnorm}[1]{\widetilde r_{\vct{\alpha}}^{(#1)}}
\newtheoremstyle{radialplain}{0pt}{0pt}{\normalfont}{}%
  {\bfseries}{.}{0.5em}{}
\theoremstyle{radialplain}
\newtheorem{proposition}{Proposition}
\usepackage{tcolorbox}
\newtcolorbox{radialpropbox}{
  sharp corners,boxrule=0pt,colback=black!5,coltext=black,
  left=8pt,right=8pt,top=7pt,bottom=7pt,boxsep=0pt,
  before skip=9pt,after skip=9pt}

%% file: sections/00-abs.tex
\begin{abstract}

Training acceleration methods for diffusion transformers, such as REPA and dispersive regularization, are commonly explained by additional supervision, spatial-structure priors, or improved feature geometry. We revisit this view and uncover a striking internal change: the AdaLN-Zero residual gate $\alpha$, which directly controls how conditioned updates enter the residual stream, is substantially reorganized by these methods. This observation motivates our hypothesis that acceleration arises, at least in part, from reshaping the residual gating pathway rather than solely from the feature structures explicitly imposed by auxiliary objectives. Supporting this view, accelerated models are markedly more sensitive to perturbations of $\alpha$, and their gate geometry reflects semantic relationships among classes. We further introduce the normalized participation ratio (nPR) to characterize the channel-wise organization of the gates and find that, across REPA teachers, projector architectures, and radial-objective targets, broader gate participation accompanies better generation quality. Guided by this insight, we propose a simple teacher-free radial objective that promotes broader gate participation through feature-norm regularization, without pairwise feature comparisons. On ImageNet $256\times256$ with SiT-B/2, our method improves FID over dispersive loss while increasing mean gate participation over timesteps and classes, further supporting the link between gate organization and training acceleration. Our results identify residual-gate reorganization as a measurable axis of diffusion training acceleration and suggest a simple direction for further acceleration.
Our code is available at \url{https://github.com/naver-ai/radialobj}
\end{abstract}

%% file: sections/01-intro.tex
\section{Introduction}
\label{sec:introduction}

Recent advances have substantially accelerated diffusion transformer training by injecting additional auxiliary objectives through representation alignment (REPA) with frozen-teacher supervision~\citep{YuEtAl2025}, spatially structured representation alignment~\citep{SinghEtAl2026}, and feature-geometry regularization~\citep{WangEtAl2025a}. Despite their differences, these approaches are largely understood through the representational structures explicitly imposed by their objectives. Yet, \textit{are these structures themselves responsible for faster training, or do they reshape a more fundamental mechanism within the transformer?}

We revisit this question through the conditioning pathway of AdaLN-Zero~\citep{PeeblesEtAl2022}. In diffusion transformers such as DiT~\citep{PeeblesEtAl2022} and SiT~\citep{MaEtAl2024}, timestep and class embeddings generate modulation vectors for every transformer block. Among them, the residual gate $\vct{\alpha}$\footnote{We refer to $\vct{\alpha}$ as a \textit{gate}, as it modulates the embeddings through channel-wise multiplication.} plays a particularly direct role: it scales each channel of a branch output before residual addition, thus controlling how strongly each feature channel contributes to the residual stream across all token positions. This suggests a complementary view of diffusion acceleration: auxiliary objectives may not only reshape the features they supervise, but also reorganize \emph{how those features are injected into the residual stream}. We therefore conjecture that residual gating may provide a common internal signature of accelerated diffusion training (\cref{fig:teaser}).

To this end, we begin with two simple interventions: removing the AdaLN-Zero modulation pathway during training and perturbing the residual gate itself at test time. Gating matters even for the baseline (without acceleration), but strikingly, accelerated models are far more vulnerable to its disruption. This contrast suggests that acceleration does more than improve the features being learned: it fundamentally changes how the model uses them through the gating pathway. Motivated by this observation, we first ask whether residual gates carry interpretable semantic structure. Using WordNet~\citep{Fellbaum1998}, we find that distances between class-conditioned gates increase with semantic distance between classes, indicating that the gating pathway reflects the semantic organization of the generated categories. Next, we quantify gate participation using the \textit{normalized participation ratio} (nPR), a scale-invariant measure of how broadly gate magnitude is distributed across channels that also guides our subsequent method design. Higher gate participation, averaged over timesteps and classes, tracks better generation quality across REPA teachers and, for a fixed teacher, across projector designs, identifying gate participation as a correlate of generation quality within each sweep.

Based on gate participation, we introduce the \ourmethod, a simple teacher-free objective that acts only on hidden-feature norms. Unlike representation alignment, it requires no pretrained teacher guidance; unlike dispersive regularization, it requires no pairwise feature comparisons. Instead, the \ourmethod penalizes deviations of each sample's flattened feature norm from a target. Despite acting only on feature norms, this objective increases residual-gate participation and improves generation quality. As a result, our method improves both Fr\'echet inception distance~\citep[FID;][]{HeuselEtAl2017} and mean gate participation, supporting residual-gate participation as a candidate mechanism underlying diffusion training acceleration. Quantitatively, on ImageNet $256\times256$~\citep{DengEtAl2009} with SiT-B/2 trained for 400K steps, the \ourmethod achieves an FID of 11.83, improving over Vanilla\footnote{We refer to Vanilla as a baseline SiT~\citep{MaEtAl2024} model.} (16.24) and Dispersive (13.19), while increasing the mean final-block MLP gate participation from 0.434 in Vanilla to 0.642. 

Overall, our results point beyond the representations explicitly targeted by existing acceleration methods. These objectives also reorganize the residual gating pathway, whose channel participation tracks generation quality in controlled sweeps. We argue that this perspective identifies residual-gate organization as a previously overlooked axis of diffusion acceleration and provides a simple route to promote it without either a representation teacher or pairwise feature comparisons.

%% file: sections/02-prelim.tex

\begin{figure}[t]
    \centering
    \includegraphics[width=.90\linewidth]{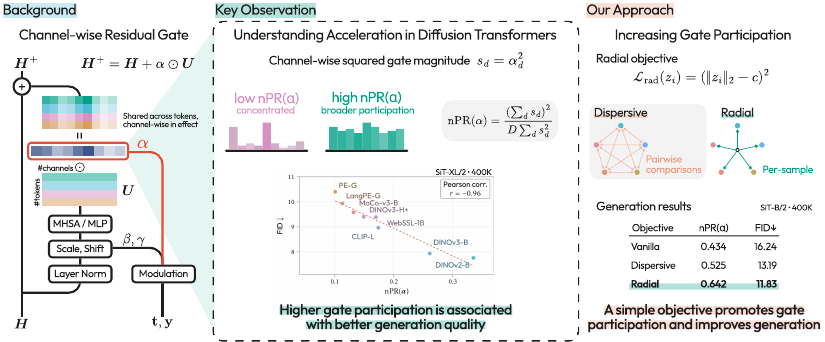}
    \caption{\textbf{Overview: from residual gate insights to improved diffusion training.}  
\textbf{\figleft} AdaLN-Zero gates scale residual-branch outputs channel-wise, with each gate shared across token positions.
\textbf{\figcenter} Our proposed \textit{normalized participation ratio (nPR)} measures how evenly squared gate magnitudes are distributed across channels. Across REPA teachers, higher mean gate participation is associated with lower FID.
\textbf{\figright} Our proposed \textit{Radial objective} penalizes per-sample feature norms without a pretrained teacher or pairwise feature comparisons. 
}
    \label{fig:teaser}
\end{figure}

\vspace{-0.5em}
\section{Background}
\vspace{-0.5em}
\label{sec:preliminaries}
\paragraph{Diffusion transformers} process latent image tokens through transformer blocks containing a multihead self-attention (MHSA) branch and an MLP branch~\citep{PeeblesEtAl2022}.
Each branch adds an update to the residual stream.
We study the AdaLN-Zero conditioning used in these blocks.

Let $\mat{H}\in\mathbb R^{N\times D}$ denote $N$ tokens with $D$ channels. 
The conditioning pathway produces the modulation vectors $\vct{\alpha},\vct{\beta},\vct{\gamma}\in\mathbb R^D$ and the output embedding $\mat{H}^+$ is sequentially computed as 
$
      \mat{H}^+=\mat{H}+\vct{\alpha}\odot\mat U
$,
where $\mat U$ can be either $\texttt{MHSA}(\widetilde{\mat H})$ or $\texttt{MLP}(\widetilde{\mat H})$, and $\widetilde{\mat{H}}= (\vct{1}+\vct{\gamma})\odot\operatorname{\texttt{LayerNorm}}(\mat{H})+\vct{\beta}$.
Here, $\vct{\gamma}$ and $\vct{\beta}$ specify the scale and shift of the normalized branch input; $\vct{\alpha}$ scales the branch output before residual addition. Accordingly, $\vct{\alpha}$ may be interpreted as a residual gating vector.
The Hadamard product $\odot$ acts channel-wise, with each vector broadcast across tokens; MHSA and MLP branches have separate modulation parameters in each block. Note that $\vct\alpha$ is structurally distinct: Whereas $\vct\beta$ and $\vct\gamma$ modulate the branch input and are mediated by the subsequent projection layers in MHSA or MLP, $\vct\alpha$ directly gates the branch output before residual addition. We thus conjecture that training acceleration is governed primarily by $\vct\alpha$.

We refer to $\vct{\alpha}$ as the \textit{channel-wise residual gate}: $\alpha_d$ multiplies channel $d$ at every token position.
Its entries are signed scaling coefficients.
The gate is obtained directly from the conditioning pathway, without an image forward pass. 
Because $\vct{\alpha}$ directly scales the residual branch output, it lets us examine how conditioning controls each channel's contribution to the residual stream.
We first test how generation quality responds to changes in the gate's channel-wise magnitude distribution, then examine the structure of gates learned under accelerated training.
Prior work has examined AdaLN initialization and parameter updates~\citep{ZhuEtAl2026} and the semantic structure of conditioning embeddings~\citep{PhamEtAl2026}.
How the residual gate values organize channel contributions under accelerated training remains underexplored.

\noindent\textbf{Diffusion transformer acceleration and its mechanics.} 
Recent auxiliary losses to accelerate diffusion transformer training fall into two broad categories: teacher-guided (REPA~\citep{YuEtAl2025} and iREPA~\citep{SinghEtAl2026}) and teacher-free (dispersive loss~\citep{WangEtAl2025a}).
REPA and iREPA accelerate training by aligning diffusion features with representations from a frozen vision foundation model through a learnable projector. Dispersive loss is a regularizer that encourages larger pairwise $\ell_2$ distances between the hidden features of samples within a batch.

%% file: sections/03-observations.tex
\section{Understanding Acceleration in Diffusion Transformers}

\subsection{Motivation}
\noindent\textbf{\hlg{What lies beyond existing representation-level supervision?}}
Existing explanations for diffusion acceleration largely focus on the representations explicitly targeted by the auxiliary objective: REPA guides hidden features toward pretrained visual representations~\citep{YuEtAl2025}, while dispersive objectives reshape their geometry~\citep{WangEtAl2025a}. We conjecture that these feature-level effects are not the full story. Instead, acceleration may critically depend on how such supervision reorganizes the AdaLN-Zero gating pathway (beyond prior initialization analyses~\citep{ZhuEtAl2026}), which directly controls the contribution of conditioned residual updates throughout the network.




\noindent\textbf{\hlg{Pilot study 1: How does gating affect generation quality during training?}}
\label{sec:gate-pilot1}
We perform a pilot study by training models with in-context conditioning~\citep{PeeblesEtAl2022} in place of AdaLN-Zero, removing its conditional modulation pathway, including the residual gate.
The FID improvements from both Dispersive and REPA become substantially smaller under in-context conditioning (\cref{tab:conditioning-mechanism}).
Relative to Vanilla, REPA's FID change shifts from $-18.24$ to $-13.48$, while Dispersive's shifts from $-4.98$ to $+1.15$.
This comparison supports a role for the AdaLN-Zero modulation pathway in the gains from auxiliary objectives; Pilot 2 isolates the residual gate.

\begin{table}[!htbp]
    \caption{\textbf{FID changes under different conditioning architectures}. Replacing AdaLN-Zero with in-context conditioning (\ie, removing the AdaLN-Zero modulation, including residual gates) substantially weakens the acceleration gains. All models are SiT-B/2 trained for 100K steps; FID ($\downarrow$) follows \cref{sec:fid-protocol}, and $\Delta$ is the change from Vanilla with the same conditioning.}
    \label{tab:conditioning-mechanism}
    \centering
    \small
    \begin{tabular}{@{}cc|@{\hspace{18pt}}cc|@{\hspace{18pt}}cc@{}}
        \toprule
        \begin{tabular}[c]{@{}c@{}}Conditioning\\architecture\end{tabular} & Vanilla & REPA & $\Delta$ & Dispersive & $\Delta$ \\
        \midrule
        AdaLN-Zero {\scriptsize (baseline)} & 45.41 & 27.17 & \textcolor{Green}{$-18.24$ ($-40.2\%$)} & 40.43 & \textcolor{Green}{$-4.98$ ($-11.0\%$)} \\
        In-context & 47.23 & 33.76 & \textcolor{Green}{$-13.48$ ($-28.5\%$)} & 48.39 & \textcolor{Red}{$+1.15$ ($+2.4\%$)} \\
        \bottomrule
    \end{tabular}
\end{table}

\noindent\textbf{\hlg{Pilot study 2: Effect of gating at test time.}}
\label{sec:gate-pilot2}
After training, we perturb the gate values of frozen Vanilla, Dispersive, and REPA models during sampling to measure their sensitivity to changes in the learned gates. This intervention evaluates sensitivity without allowing the model weights to adapt to the modified gates. We use SiT-XL/2 checkpoints trained for 400K steps. 
At each sampling step, we redistribute the squared magnitudes of the MLP gate at the eighth layer (L8), where the REPA objective is applied, making the distribution either more uniform or more concentrated while preserving the gate’s signs and $\ell_2$ norm.
We match the relative perturbation magnitude $\delta=\|\vct{\alpha}^{\textrm{new}}-\vct{\alpha}\|_2/\|\vct{\alpha}\|_2$ across these directions and models.
\begin{figure}[t]
    \centering
    \includegraphics[width=.95\linewidth]{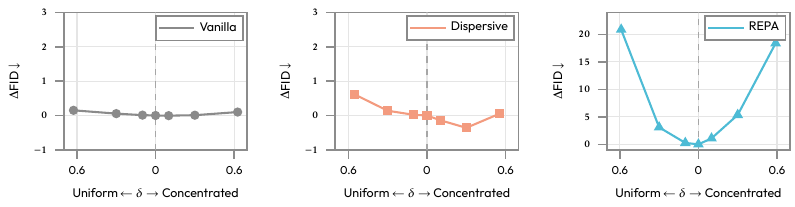}
    \vspace{-.5em}
    \caption{\textbf{Does gate redistribution in Vanilla, Dispersive, and REPA affect FID?} The center (\ie, $x=0$) is the unmodified baseline; moving left makes gate magnitudes more uniform, and moving right makes their distribution more concentrated. Tick labels give relative gate perturbation magnitude $\delta$. All three panels show FID change from each model's own baseline. 
    }
    \label{fig:gate-intervention}
\end{figure}
At a matched perturbation magnitude of $\delta=0.30$, Dispersive shows larger absolute FID changes than Vanilla, while REPA responds much more strongly (\cref{fig:gate-intervention}).
Both studies show that generation quality is sensitive to $\vct\alpha$, most strongly in REPA ($\Delta\text{FID}$ up to $+5.356$ at $\delta=0.30$).
With frozen weights, this test measures sensitivity to departures from the learned gates, not a preference for uniform gates; \cref{sec:npr-quality} compares gates learned under different training.
Since REPA/iREPA is primarily understood through representation alignment, this raises a question: \textit{Could the gating pathway be an underlying piece of the acceleration?}


\begin{figure}[b]
\begin{minipage}{0.37\linewidth}
    \centering
    \includegraphics[width=.9\linewidth]{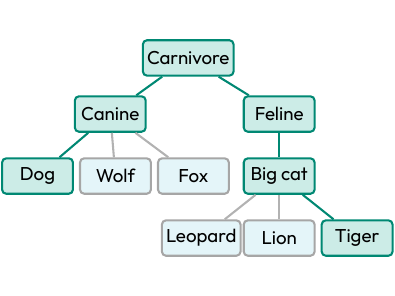}
    \caption{\textbf{Illustration of a WordNet hierarchy and path distance.} The highlighted shortest path from dog to tiger spans five hypernym edges.}
    \label{fig:wordnet-path}
\end{minipage}
\hfill
\begin{minipage}{0.6\linewidth}
    \centering
    \includegraphics[width=.95\linewidth]{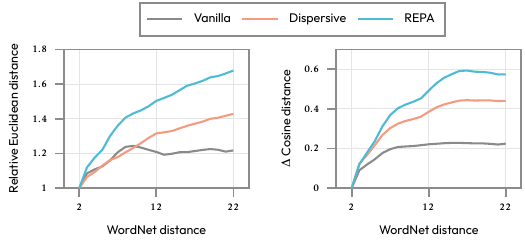}
    \caption{\textbf{Change in residual gates versus WordNet distance}. We compare how the gates change in Vanilla, REPA, and Dispersive models at $t=0.5$. \textbf{\figleft} Euclidean distance between gate magnitudes $\vct{a}_i=|\vct{\alpha}_i|$, divided by its bin mean at WordNet distance 2. \textbf{\figright} Cosine distance after centering $\vct{a}_i$ across classes, shifted by its bin mean at WordNet distance 2 to start at zero.}
    \label{fig:gate-wordnet-distance}
\end{minipage}
\end{figure}

\subsection{Empirical Insights with Residual Gating}
Here, we investigate the role of the residual gate in accelerated diffusion-transformer training. Our pilot study revealed that generation quality in existing training acceleration methods is unexpectedly sensitive to the residual gate $\vct{\alpha}$, which suggests that the gating pathway may play a more central role than previously appreciated. This leads us to revisit REPA's acceleration from a different perspective. We study the structure of the learned gates to explore this possibility.    
\subsubsection{How Do Residual Gates Behave in Training Acceleration Methods?}
\begin{figure}[b]
    \centering
    \plotfigure{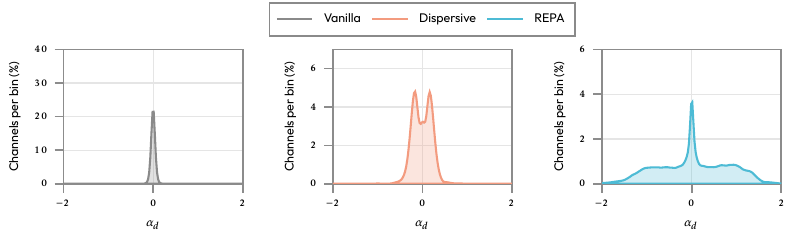}
    \caption{Histograms of signed learned MLP gate entries $\alpha_d$ for Vanilla, REPA, and Dispersive, using the same models as \cref{fig:gate-wordnet-distance}. At $t=0.5$, we pool all 1,000 classes and 1,152 channels without normalization, using common width-0.025 bins, lightly smoothed with a Gaussian kernel ($\sigma=1$ bin). All three panels show $[-2,2]$ with independently scaled linear y-axes. Bin percentages use all entries, including those outside the displayed range.}
    \label{fig:gate-channel-histogram}
\end{figure}

\noindent\textbf{\hlg{Residual gates reflect semantic distance.}} We examine how class-conditioned gate distances respond to semantic distance, and whether this response differs across training objectives.
The semantic distance is measured as WordNet distance~\citep{Fellbaum1998}, which is the number of edges on a shortest path between class synsets.
The partial WordNet noun hierarchy in \cref{fig:wordnet-path} highlights the five-edge path from dog to tiger.
We compare the class-conditioned MLP gates of Vanilla, REPA, and Dispersive; experimental settings are provided in \cref{sec:gate-structure-details}.
We use $t=0.5$ as a representative midpoint of the diffusion trajectory; \cref{sec:gate-timestep-checks} reports the same analysis at $t=0.25$ and $0.75$.
We compute distances for all 499,500 unordered class pairs and average them at each WordNet distance.
Every curve attains its minimum bin mean at WordNet distance 2.
Because overall gate scale differs across models (\cref{fig:gate-channel-histogram}), we divide each Euclidean curve by this minimum; for the scale-invariant centered cosine distance, we subtract it.
Gate distances generally increase with semantic distance in all three models, with the largest increase under REPA (\cref{fig:gate-wordnet-distance}).
At WordNet distance 22, Euclidean distances reach $1.678\times$ their starting value for REPA, compared with $1.217\times$ for Vanilla and $1.428\times$ for Dispersive; centered cosine distances increase by $0.574$, $0.225$, and $0.441$, respectively.


The hierarchy analysis shows that class-conditioned gates reflect semantic hierarchy, as the distance between gate vectors increases with semantic distance between classes. Since a larger distance between two gate vectors indicates greater separation in gate space, this trend suggests that REPA induces the strongest semantic separation among class-conditioned gates.
Together, these observations motivate quantifying the channel-wise magnitude distributions of these gates and examining their relationship with generation quality across training configurations.

\subsubsection{Defining Gate Participation and Its Impact}
\noindent\textbf{\hlg{Defining gate participation metric.}} 
We quantify gate participation with the participation ratio (PR), which measures how evenly the squared gate magnitudes are distributed across channels.
For a nonzero gate vector with squared magnitudes $s_d=\alpha_d^2$, the PR and its normalized form (nPR)\footnote{\citet{PhamEtAl2026} also employ nPR to analyze the semantic structure of conditioning embeddings. We instead measure it on the residual gates and compare it across training objectives and against generation quality.} are defined as follows:
\begin{equation}
    \mathrm{PR}(\vct{\alpha}) = \frac{\left(\sum_{d=1}^D s_d\right)^2}{\sum_{d=1}^D s_d^2},
    \quad
    \mathrm{nPR}(\vct{\alpha}) = \frac{\mathrm{PR}(\vct{\alpha})}{D},
    \quad
    \vct{\alpha} \in \mathbb{R}^D.
\end{equation}
We refer to \ourmetric as gate participation. Higher \ourmetric indicates that squared gate magnitude is spread more evenly across channels; lower \ourmetric indicates concentration in fewer channels.
Equal nonzero magnitudes on $m$ of the $D$ channels give $\mathrm{nPR}(\vct{\alpha})=m/D$, with a range of $[1/D,1]$ for nonzero gates.
The statistic is invariant to uniform nonzero rescaling.
It measures the allocation of gate magnitude across channels, rather than the overall gate scale of the residual update.


We then inspect the learned MLP gate values in Vanilla, REPA, and Dispersive. \cref{fig:gate-channel-histogram} pools the signed gate entries across all 1,000 classes at $t=0.5$, without rescaling them.
Their overall scales differ, with standard deviations of 0.049 for Vanilla, 0.285 for Dispersive, and 0.793 for REPA.

\begin{figure}[t]
    \centering
    \begin{subfigure}[t]{0.32\textwidth}
        \centering
        \includegraphics[width=\linewidth]{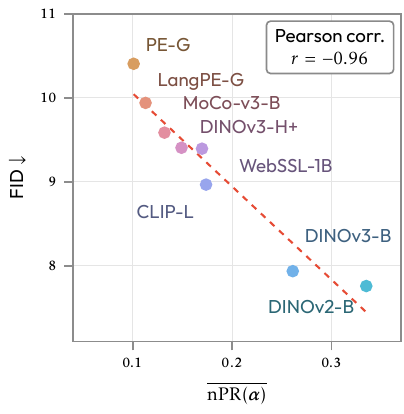}
        \vspace{-2em}
        \caption{Teacher sweep}
        \label{fig:repa-teacher-sweep}
    \end{subfigure}\hfill%
    \begin{subfigure}[t]{0.32\textwidth}
        \centering
        \includegraphics[width=\linewidth]{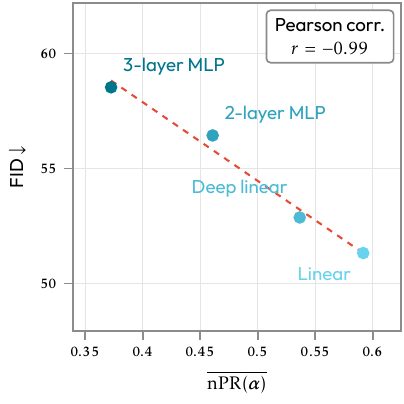}
        \vspace{-2em}
        \caption{Projector sweep}
        \label{fig:dinov2-projector-sweep}
    \end{subfigure}\hfill%
    \begin{subfigure}[t]{0.32\textwidth}
        \centering
        \includegraphics[width=\linewidth]{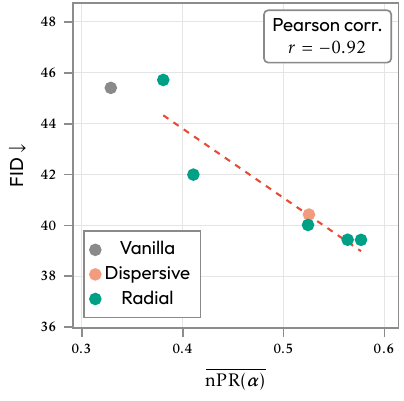}
        \vspace{-2em}
        \caption{Radial objective}
        \label{fig:norm-only-c-sweep}
    \end{subfigure}
    \caption{\textbf{Gate participation and generation quality across teacher, projector, and radial-target sweeps}. \captiona REPA with eight teachers. \captionb Projector architectures under a fixed teacher. \captionc Radial-target sweep at 100K steps, with Vanilla and Dispersive as reference points. All three horizontal axes show \ourmetricmean over 101 uniformly spaced model timesteps in $[0,1]$ and 1,000 classes. Fits and correlations use the sweep points only, excluding the reference points in \captionc.}
    \label{fig:participation-sweeps}
\end{figure}

\noindent\textbf{\hlg{Gate participation correlates with generation quality.}}
\label{sec:npr-quality}
We next examine whether mean gate participation over timesteps and classes is actually correlated with generation quality. We vary the representation-alignment target to examine the relationship 
across training configurations.
A trainable projector maps the diffusion features to the teacher representation, with alignment applied at corresponding tokens. We conduct class-conditional ImageNet $256\times256$ generation and report FID. 
Throughout, we measure gate participation at the block where the auxiliary objective is applied: L8 for SiT-XL/2 REPA, L4 for SiT-B/2 REPA, and L12 for Dispersive and the \ourmethod, with Vanilla references measured at the same block.
We measure MLP gates at L8 in the released REPA checkpoints~\citep{YuEtAl2025} corresponding to eight teacher encoders.
For each checkpoint, we compute $\mathrm{nPR}(\vct{\alpha}(t,y))$ separately for 1,000 classes and 101 uniformly spaced timesteps, then average the 101,000 values.
Across eight REPA teachers, mean MLP gate participation at L8 is strongly negatively correlated with FID ($r=-0.962$; \cref{fig:repa-teacher-sweep}).
The teacher sweep reveals an association between gate participation and generation quality across alignment targets.


\noindent\textbf{\hlg{Gate participation can track quality with a fixed teacher.}} 
\label{sec:projector-sweep} 
We now verify that the association between mean gate participation and generation quality also holds when the teacher is fixed and the projector changes; fixing the teacher removes teacher quality as a common cause of both gate participation and FID.
iREPA~\citep{SinghEtAl2026} replaces REPA's three-layer MLP projector with a $3\times3$ convolution, providing a precedent for varying the projection head.
\cref{tab:projector-components} compares four projector architectures under fixed teacher, target preprocessing, alignment weight, and training budget.

\begin{table}[!htbp]
    \caption{\textbf{Projector components and results.} We use SiT-B/2 trained for 50K steps. Check marks and crosses indicate present and absent components; $\mat{W}_i$ denotes affine slots. \ourmetricmean is computed for MLP gates at the fourth layer (L4) over 101 uniformly spaced model timesteps in $[0,1]$ and 1,000 classes. All four use the same teacher targets and alignment settings (\cref{sec:projector-sweep-details}).}
    \label{tab:projector-components}
    \centering
    \small
    \setlength{\tabcolsep}{5pt}
    \newcommand{\componentyes}{\color{Green}\ding{51}}
    \newcommand{\componentno}{\color{Red}\ding{55}}
    \begin{tabular}{@{}lccccccr@{}}
        \toprule
        Projector & $\mat{W}_1$ & SiLU & $\mat{W}_2$ & SiLU & $\mat{W}_3$ & \ourmetricmean & FID $\downarrow$ \\
        \midrule
        3-layer MLP & \componentyes & \componentyes & \componentyes & \componentyes & \componentyes & 0.373 & 58.53 \\
        2-layer MLP & \componentyes & \componentyes & \componentno & \componentno & \componentyes & 0.461 & 56.44 \\
        Deep linear & \componentyes & \componentno & \componentyes & \componentno & \componentyes & 0.536 & 52.87 \\
        Linear ($1\times1$ conv.) & \componentyes & \componentno & \componentno & \componentno & \componentno & 0.592 & \textbf{51.33} \\
        \bottomrule
    \end{tabular}
\end{table}


Reducing the MLP from three affine layers to two and then to one increases mean gate participation from $0.373$ to $0.461$ and $0.592$, while lowering FID from $58.53$ to $56.44$ and $51.33$.
Removing the nonlinearities while retaining three affine layers gives FID $52.87$.
Across the four variants, L4 MLP gate participation averaged over timesteps and classes correlates with FID at $r=-0.991$ (\cref{fig:dinov2-projector-sweep}).
Together with the teacher sweep, these results show that the participation-quality association appears both when the alignment target changes and when the projector changes under a fixed teacher.
This raises the question of whether encouraging higher gate participation improves generation quality.

%% file: sections/04-action.tex
\vspace{-.5em}
\section{Increasing Gate Participation for Better Generation Quality}
\label{sec:increasing-participation}
\vspace{-.5em}
This section studies whether encouraging higher gate participation during training improves generation quality. 
We introduce the \ourmethod, which acts on hidden-feature norms, and examine when its local gradient contribution increases gate participation.

\noindent\textbf{\hlr{The radial objective.}} 
Based on our empirical studies above, we propose a simple training acceleration method, coined \ourmethod, which adds a per-sample feature-norm penalty to the velocity prediction loss~\citep{MaEtAl2024}. Our insight is that the embedding right before the residual addition directly interacts with the residual gate $\vct\alpha$, which would likely increase gate participation (\ie, \ourmetric) and ultimately accelerate training.
Given the output $\mat{Z}_i\in\mathbb R^{N\times D}$ of the final block, define $\vct{z}_i=\operatorname{flatten}(\mat{Z}_i)$ and optimize the following objective:
\begin{equation}
    \mathcal L=\mathcal L_{\mathrm{velocity}}
    +\lambda\,\mathbb E_i\big[(\|\vct{z}_i\|_2-c)^2\big],
    \label{eq:norm-only}
\end{equation}
where the norm is over all tokens and channels of one sample, and $c$ denotes a flattened feature-norm target; $\lambda>0$ weights the auxiliary penalty.
We sweep the radial target over $c\approx k\sqrt{ND}$ with $k\in\{2,4,8,10,14\}$ and report the results for each target; \cref{sec:radial-target-choice} gives the exact values. We use the target with the lowest observed FID ($k=10$) for the performance comparison.
This objective requires neither a teacher nor distances between different samples.

\noindent\textbf{\hlr{Promoting gate participation with our radial objective.}} 
At the current model state, fix a condition $(t,y)$ and analyze its gate
$\vct{\alpha}=\vct{\alpha}_\theta(t,y)$. For an input at this condition,
the final-block output is
$\mat{Z}=\mat{H}+\vct{\alpha}\odot\mat{U}=\mat{H}+\mat{U}\operatorname{diag}(\vct{\alpha})$.
Below, $\mathbb E[\,\cdot\mid t,y]$ averages over images and noises at
the same $(t,y)$; it does not require an actual minibatch to repeat $(t,y)$.
To isolate gate-weighted channel norms, we drop the
skip--residual cross term from $\|\mat{H}+\mat{U}\operatorname{diag}(\vct{\alpha})\|_{\mathrm F}^2$
and analyze the norm
\begin{equation}\label{eq:radial-surrogate}
 \widetilde r_{\vct{\alpha}}
 =\left(
  \|\mat{H}\|_{\mathrm F}^{2}
  +\sum_{d=1}^{D}\alpha_d^{2}\|\mat{U}_{:,d}\|_2^{2}
 \right)^{1/2},
\end{equation}
where $\|\mat{U}_{:,d}\|_2^2$ is the squared pre-gating norm of channel $d$, summed
over tokens. We denote the radial objective evaluated with this norm,
$\mathbb E[(\widetilde r_{\vct{\alpha}}-c)^2\mid t,y]$, by $\radialsurrogate$.
We assume \emph{channel-symmetric} pre-gating norm patterns within this
conditional analysis, assigning no a priori preference to any residual channel.
This assumption puts channels on equal footing before gating
(\cref{sec:radial-gate-analysis}).

\begin{radialpropbox}
\begin{proposition}[Gate participation under channel symmetry]
\label{prop:radial-channel-symmetry}
Under channel symmetry and the regularity conditions in
\cref{sec:radial-gate-analysis}, a sufficiently small conditional gate update
$\vct{\alpha}^+=\vct{\alpha}-\eta\nabla_{\vct{\alpha}}\radialsurrogate(\vct{\alpha})$, with $\eta>0$,
satisfies
\begin{equation}\label{eq:radial-participation-nondecrease}
 \mathrm{nPR}(\vct{\alpha}^+)\ge\mathrm{nPR}(\vct{\alpha}).
\end{equation}
\end{proposition}
\end{radialpropbox}

The mechanism follows from the relative gate gradient,
\begin{equation}\label{eq:radial-relative-update}
 -\frac{1}{\alpha_d}\frac{\partial\radialsurrogate}{\partial\alpha_d}
 =2\,\mathbb E\!\left[
 \left(\frac{c}{\widetilde r_{\vct{\alpha}}}-1\right)
 \|\mat{U}_{:,d}\|_2^2
 \,\middle|\,t,y\right],
\end{equation}
which equals $\Delta\alpha_d/(\eta\alpha_d)$ with $\Delta\alpha_d=\alpha_d^+-\alpha_d$. 
$c/\widetilde r_{\vct{\alpha}}-1$ is specific to each input embedding and shared by all channels of that input; channels differ only
through $\|\mat{U}_{:,d}\|_2^2$ and $\alpha_d$.
Consider two otherwise matched input norm patterns at the same $(t,y)$,
differing only by exchanging two pre-gating channel norms. If the larger channel
norm is paired with the larger gate, the whole-input norm becomes larger,
reducing the shared radial coefficient; swapping the channel norms gives the
opposite effect. Under channel symmetry, the two patterns are weighted equally.
Averaging over such realizations gives weaker gates no smaller
fractional magnitude updates than stronger ones:
\begin{equation}\label{eq:radial-update-order}
 |\alpha_i|>|\alpha_j|
 \quad\Longrightarrow\quad
 \frac{\Delta\alpha_i}{\alpha_i}
 \le
 \frac{\Delta\alpha_j}{\alpha_j}.
\end{equation}
For sufficiently small, sign-preserving steps, weaker gates can thus grow
faster or shrink more slowly, reducing gate concentration.
Under the same analysis, positive and negative quadratic norm objectives yield
uniform relative gate rescaling, whereas the radial objective introduces
input-dependent inverse-norm weighting.
This result concerns the conditional gate-space update in the
cross-term-excluded analysis, rather than the full parameter update during
training.


\noindent\textbf{\hlr{Higher participation and lower FID under radial training.}}
We search over target norms using SiT-B/2 trained for 100K steps.
We calibrate $\lambda$ over the first five training iterations so that the weighted auxiliary gradient norm is approximately 10\% of the velocity-loss gradient norm.
We measure MLP gates at the twelfth layer (L12), on which the dispersive loss and the radial objective act.
The target is selected by the lowest observed FID; \cref{sec:radial-target-choice} reports the search values and sensitivity. Across the five settings in \cref{fig:norm-only-c-sweep}, L12 MLP gate participation averaged over timesteps and classes is negatively correlated with FID ($r=-0.916$).
These measurements test whether radial training yields higher participation together with improved generation quality. 
FID decreases from 45.72 at the lowest target to 39.43 at the selected target ($k=10$), while mean gate participation increases from 0.381 to 0.577.

\noindent\textbf{\hlr{Generation quality with and without a teacher.}}
\cref{tab:performance-comparison} compares SiT-B/2 models at 400K training steps.
We train \ourmethod with the selected target $c=10\sqrt{ND}$.
Its FID of 11.83 improves on Vanilla (16.24) and Dispersive (13.19) while requiring no teacher.
Its mean L12 MLP gate participation of 0.642 is likewise the highest, compared with 0.434 for Vanilla and 0.525 for Dispersive.

\begin{table}[t]
    \caption{\textbf{Generation quality at 400K training steps}, SiT-B/2 at $256\times256$. We report the results of Vanilla, Dispersive, and the proposed \ourmethod, with REPA as a teacher-guided reference. The best teacher-free results are in bold.}
    \label{tab:performance-comparison}
    \centering
    \small
    \begin{tabular}{lrrrrr}
        \toprule
        Method & FID $\downarrow$ & sFID $\downarrow$ & IS $\uparrow$ & Precision $\uparrow$ & Recall $\uparrow$ \\
        \midrule
        \multicolumn{6}{l}{\textit{teacher-free}} \\
        Vanilla~\citep{MaEtAl2024} & 16.24 & 5.43 & 86.67 & 0.670 & 0.557  \\
        Dispersive~\citep{WangEtAl2025a} & 13.19 & \textbf{5.25} & 98.19 & 0.696 & 0.553 \\
        \rowcolor{my2ndblue}
        \textbf{\ourmethod (ours)} & \textbf{11.83} & 5.27 & \textbf{105.93} & \textbf{0.714} & \textbf{0.559} \\
        \midrule
        \multicolumn{6}{l}{\teacherentry{\textit{teacher-guided}}} \\
        \teacherentry{REPA~\citep{YuEtAl2025}} & \teacherentry{7.71} & \teacherentry{5.40} & \teacherentry{141.54} & \teacherentry{0.735} & \teacherentry{0.577} \\
        \bottomrule
    \end{tabular}
    \vspace{-.5em}
\end{table}

\begin{figure}[t]
    \centering
    \begin{subfigure}[t]{0.33\textwidth}
        \centering
        \makebox[\linewidth][c]{\includegraphics{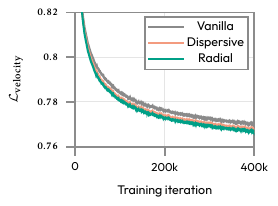}}
    \end{subfigure}\hfill%
    \begin{subfigure}[t]{0.33\textwidth}
        \centering
        \makebox[\linewidth][c]{\includegraphics{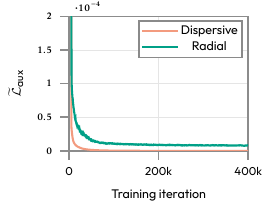}}
    \end{subfigure}\hfill%
    \begin{subfigure}[t]{0.33\textwidth}
        \centering
        \makebox[\linewidth][c]{\includegraphics{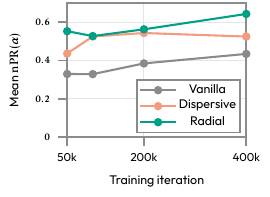}}
    \end{subfigure}
    \vspace{-.5em}
    \caption{\textbf{Training dynamics over 400K steps.} \textbf{\figleft} Training MSE. \textbf{\figcenter} Normalized auxiliary losses. \textbf{\figright} Mean \ourmetric over 101 uniformly spaced timesteps in $[0,1]$ and 1,000 classes.}
    \label{fig:radial-gate-trajectories}
    \label{fig:training-losses}
\end{figure}

\noindent\textbf{\hlr{Gate participation during training.}} 
We track gate participation at 50K, 100K, 200K, and 400K training steps for Vanilla, Dispersive, and Radial.
At each checkpoint, we evaluate L12 MLP gate participation averaged over timesteps and classes using EMA weights;
\cref{fig:radial-gate-trajectories} (right) connects these measurements without smoothing.
Radial maintains higher gate participation than Vanilla at all four checkpoints.
Dispersive also exceeds Vanilla, with participation peaking at 200K.

\noindent\textbf{\hlr{Training-loss visualization.}} 
We also report the recorded velocity-prediction MSE and auxiliary training losses. Here, $\mathcal L_{\mathrm{aux}}$ denotes the additional regularization term: the dispersive loss for Dispersive and the feature-norm penalty in \cref{eq:norm-only} for Radial, each measured before weighting by its coefficient. Vanilla has no auxiliary term.
For the auxiliary curves in \cref{fig:training-losses} (center), we plot the normalized gap $\widetilde{\mathcal L}_{\mathrm{aux}}=(L_t-L_{\min})/(L_0-L_{\min})$, where $L_t$ is the raw auxiliary loss at training step $t$, $L_0$ is its step-zero value, and $L_{\min}=-\ln 32$ for Dispersive and $0$ for Radial.

%% file: sections/05-discussion.tex
\section{Discussion and Limitations}
\label{sec:discussion}

\noindent\textbf{Interpreting iREPA and spatial structure.}
iREPA motivates its convolutional projector and spatial normalization as ways to improve the transfer of spatial structure from teacher representations~\citep{SinghEtAl2026}.
Its spatial diagnostics, including LDS, summarize how similarity between patch tokens varies with their spatial separation.
Our fixed-teacher projector sweep in \cref{sec:projector-sweep} helps distinguish properties that change together when replacing an MLP with a convolution: nonlinear depth, channel transformation, and spatial mixing.
At 50K steps, the $1\times1$ projector reaches FID 51.33, compared with 51.68 for an additional $3\times3$ convolutional projector trained under the same teacher normalization and alignment settings.
The comparable quality of these two heads shows that neighboring-token mixing inside the projector is not required to attain this performance in the tested configuration.
The lower FID of the deep linear head relative to the nonlinear MLP also motivates examining the effect of projector nonlinearities.



\noindent\textbf{Association and intervention.}
The teacher and projector sweeps show a recurring relationship between gate participation and FID within their respective experimental settings.
The post-hoc pilot establishes sensitivity to gate modification, but its uniformizing interventions do not consistently improve generation quality.
The WordNet analysis shows class-related structure in the gates; it does not establish a lower bound on the participation needed to represent a semantic hierarchy.
The \ourmethod provides a simple teacher-free training objective that yields higher participation and lower FID in the tested target sweep.
It acts on feature norms and affects the gates indirectly, alongside other model parameters.
The local gradient proposition specifies conditions for a positive radial contribution (\cref{sec:radial-surrogate-scope}); it does not guarantee participation growth under the full optimizer or establish that such growth causes the FID improvement.


%% file: sections/05-rel.tex
\section{Related Work}

\noindent\textbf{Representation alignment in diffusion transformers.}
REPA supervises intermediate diffusion features with pretrained visual representations, motivated by the difficulty of learning useful representations from the generative objective alone~\citep{YuEtAl2025}.
Subsequent work examines the alignment target and projector in iREPA~\citep{SinghEtAl2026}, and the timing and scope of alignment in HASTE, which combines feature and attention-map supervision before terminating alignment during training~\citep{WangEtAl2025b}.
Our analysis concerns the internal modulation induced by these training choices, focusing on residual-gate participation and geometry.

\noindent\textbf{Spatial structure in diffusion representations.}
iREPA compares global semantic performance with spatial self-similarity in teacher representations and motivates changes intended to improve the transfer of spatial information~\citep{SinghEtAl2026}.
HASTE also incorporates spatial relationships through attention-map supervision~\citep{WangEtAl2025b}.
Our projector experiments instead examine the role of projection architecture within the tested alignment configurations, without making a broader claim about the necessity of spatial information in diffusion models. 

\noindent\textbf{Adaptive modulation and gating in DiTs.}
DiT uses adaptive normalization and residual scaling through AdaLN-Zero~\citep{PeeblesEtAl2022}; SiT adopts this transformer backbone for interpolant-based generative modeling~\citep{MaEtAl2024}.
Related analyses examine modulation initialization~\citep{ZhuEtAl2026} and the concentration of semantic information in conditional embeddings~\citep{PhamEtAl2026}.
We focus specifically on the learned branch-specific residual gate $\vct{\alpha}$, its allocation across channels, and its geometry across training objectives.

\noindent\textbf{Representation geometry and effective dimensionality.}
Sparsity measures characterize the concentration of coefficients~\citep{HurleyEtAl2008}, while intrinsic-dimension estimates have been used to assess self-supervised representations~\citep{MordacqEtAl2026}.
Dispersive training modifies representation geometry through an auxiliary separation objective~\citep{WangEtAl2025a}.
We apply participation ratio to squared gate coordinates in the model's channel basis.
This measures the allocation of gate magnitude across channels, rather than the spectral rank or intrinsic dimension of hidden representations, and serves here as a diagnostic associated with generation quality.

%% file: sections/06-con.tex
\section{Conclusion}

We examined diffusion transformer acceleration through class-conditioned residual gates.
Accelerated models depend more strongly on these gates: removing the AdaLN-Zero modulation shrinks their gains, and perturbing $\vct{\alpha}$ degrades REPA the most.
The gates also reflect the semantic hierarchy among classes, most strongly under REPA, and their gate participation tracks FID across REPA teachers and, for a fixed teacher, across projectors.
Building on this, the teacher-free \ourmethod raised gate participation and improved FID over Dispersive without pairwise feature comparisons.
Because gate participation is computed from the conditioning pathway alone, without an image forward pass, it offers a lightweight quantity to monitor in accelerated diffusion training.

\subsection*{AI use statement}
In this work, we used generative AI tools to write small pieces of code for implementing our methods, to assist in writing proofs, and to formulate mathematical claims.
We have not used generative AI tools to develop theoretical models or conceptual frameworks, to provide critical ingredients for proofs, to propose or refine hypotheses, to design or give feedback on the research methodology or experiments, to interpret results, or to translate text.
Synthetic data generation, dataset cleaning and reformatting, and qualitative or
thematic data analysis are not applicable to this work.
Additionally, we used generative AI tools to search for related work, to edit the paper for readability, and to check its formatting.
We emphasize that generative AI did not originate the core methodology or generate the full experimental codebase.
We have reviewed all AI-assisted work: the authors derived and verified all mathematical claims and proofs themselves, AI-assisted code was validated through sanity checks and by reproducing published baseline results, and every work found through AI-assisted search was checked against the original paper.
We take responsibility for the final content of this work, including text, claims, code, analyses, and results produced with the aid of generative AI.


\subsection*{Reproducibility statement}

We provide all implementation details and hyperparameters in
\cref{sec:reproducibility,sec:gate-analysis-details,sec:radial-details}.

%% file: sections/07-reproducibility.tex
\newpage
\appendix
\renewcommand\thefigure{\Alph{figure}}
\setcounter{figure}{0}
\renewcommand\thetable{\Alph{table}}
\setcounter{table}{0}
\crefalias{section}{appendix}
\crefalias{subsection}{appendix}

\section*{\Large Appendix}
\section{Common Setup}
\label{sec:reproducibility}

\subsection{Models and Training}
\label{sec:experimental-setting}
All experiments use class-conditional ImageNet~\citep{DengEtAl2009} generation at $256\times256$ resolution with SiT~\citep{MaEtAl2024} and AdaLN-Zero, except the in-context conditioning ablation of Pilot study 1 (\cref{sec:icc-pilot-details}).
The teacher sweep, gate intervention, hierarchy analysis, and gate histograms use SiT-XL/2, with 28 blocks, hidden width 1,152, and 16 attention heads.
The projector and \ourmethod experiments use SiT-B/2, with 12 blocks, width 768, and 12 heads.
Both architectures use $2\times2$ latent patches and an MLP expansion ratio of four.
Layer numbers are one-based: L1 denotes the first transformer block.

\paragraph{Local SiT-B/2 training.}
Unless stated otherwise, our SiT-B/2 runs use seed 42, global batch size 256, bfloat16 mixed precision, and AdamW~\citep{LoshchilovEtAl2019} with a constant learning rate of $10^{-4}$, $\beta_1=0.9$, $\beta_2=0.999$, and zero weight decay.
The label-drop probability is 0.1 and the EMA decay is 0.9999.
Our SiT-B/2 Dispersive runs use the best-performing PyTorch configuration of the official implementation~\citep{WangEtAl2025a}: the InfoNCE-L2 dispersive loss on the final-block output (L12) with coefficient $\lambda=0.25$ and temperature $\tau=1$, computed within each per-GPU batch of 32 on eight GPUs.
The velocity objective uses a linear interpolation path.
Training uses cached ImageNet latents from the ft-MSE VAE, sampled from the stored posterior.

\subsection{Gate Participation Measurement}
\label{sec:gate-measurement}
Unless a fixed timestep is stated, we evaluate the conditioning pathway on 101 evenly spaced timesteps in $[0,1]$, including both endpoints, and all 1,000 ImageNet class labels.
For each MLP gate vector $\vct{\alpha}(t,y)\in\mathbb{R}^D$, we compute $\mathrm{nPR}(\vct{\alpha})=(\sum_d\alpha_d^2)^2/(D\sum_d\alpha_d^4)$ and take the arithmetic mean over the 101,000 conditions, which gives \ourmetricmean.
We average participation values rather than gate vectors, and do not concatenate MHSA and MLP gates.
Gates are computed in FP32 and statistics in FP64.
We use the convention $\vct{x}_t=(1-t)\vct{x}_{\mathrm{data}}+t\vct{x}_{\mathrm{noise}}$ and convert it to each checkpoint's input convention.
In the scatter plots, regression lines are ordinary least-squares fits of FID to \ourmetricmean, and $r$ is the Pearson correlation across the plotted runs.

\subsection{FID Evaluation}
\label{sec:fid-protocol}
Local FIDs use the ADM evaluator~\citep{DhariwalEtAl2021} and the ImageNet-256 reference statistics, with 50,000 generated images, EMA model weights, the \texttt{sd-vae-ft-ema} decoder, CFG 1.5, and sampling seed 42.
The sampler is adaptive Dormand--Prince (\texttt{dopri5})~\citep{DormandEtAl1980}, with absolute tolerance $10^{-6}$ and relative tolerance $10^{-3}$.
Besides FID~\citep{HeuselEtAl2017}, the evaluator reports sFID~\citep{DhariwalEtAl2021}, inception score (IS)~\citep{SalimansEtAl2016}, and precision and recall~\citep{KynkaanniemiEtAl2019}.
The gate intervention pilot and the teacher sweep use different protocols, described in \cref{sec:gate-pilot-details,sec:teacher-sweep-details}.

\section{Details of the Gate Analysis}
\label{sec:gate-analysis-details}

\subsection{In-Context Conditioning Pilot}
\label{sec:icc-pilot-details}
\cref{tab:conditioning-mechanism} compares six SiT-B/2 runs trained for 100K steps with the setup of \cref{sec:experimental-setting} and evaluated at their 100K EMA checkpoints with the local FID protocol of \cref{sec:fid-protocol}.
Within each conditioning architecture, the three runs differ only in the auxiliary objective.
REPA aligns L4 with DINOv2-B through a three-layer MLP projector with coefficient 0.5, and Dispersive uses coefficient 0.25 at the final block (L12).

\paragraph{In-context conditioning.}
Following DiT~\citep{PeeblesEtAl2022}, we replace AdaLN-Zero with two conditioning tokens, a timestep embedding and a class embedding, prepended to the patch tokens.
The blocks are standard pre-LN transformer blocks with affine LayerNorm and no scale, shift, or gate.
The final layer is an affine LayerNorm followed by a linear projection without modulation, and the two conditioning tokens are discarded before unpatchifying.
Patch tokens use fixed 2D sine-cosine positional embeddings, and the conditioning tokens receive zero positional embeddings.
Auxiliary losses are computed on the patch tokens only.

\subsection{Gate Intervention Pilot}
\label{sec:gate-pilot-details}
The pilot uses SiT-XL/2 EMA checkpoints at 400K steps for Vanilla, Dispersive, and REPA with DINOv2-B~\citep{OquabEtAl2024}; the REPA checkpoint is the DINOv2-B checkpoint of the teacher sweep (\cref{sec:teacher-sweep-details}).
To compare Dispersive and REPA on the same block, the Dispersive checkpoint applies the dispersive loss at L8, where these REPA checkpoints apply alignment, with coefficient 0.5 and $\tau=1$.
At every sampling step, we modify only the realized L8 MLP gate.
Writing $p_d=\alpha_d^2/\|\vct{\alpha}\|_2^2$, the uniform direction uses $p'_d=(1-\epsilon)p_d+\epsilon/D$, while the concentrated direction uses $p'_d=p_d^\kappa/\sum_j p_j^\kappa$ with $\kappa>1$.
We reconstruct $\alpha'_d=\operatorname{sign}(\alpha_d)\|\vct{\alpha}\|_2\sqrt{p'_d}$ and solve for $\epsilon$ or $\kappa$ separately for each condition to match the target relative perturbation $\delta$.
This preserves signs, gate norm, and channel ordering.
The dose grid contains $\delta=0.10$, $0.30$, and a model-specific near-maximal feasible dose ($0.627$ for Vanilla, $0.555$ for Dispersive, and $0.590$ for REPA).

Each evaluation uses 50,000 generated images, CFG 1.0~\citep{HoEtAl2022}, and 50-step Euler sampling, with sampling seed 42 shared between interventions and their unmodified baselines.
Vanilla and Dispersive use the SiT sampling convention and ft-EMA decoder~\citep{RombachEtAl2022}, while the REPA checkpoint uses the REPA convention and ft-MSE decoder.
Because the pipelines differ, we report FID changes relative to each model's own baseline.

\subsection{Gate Histograms and WordNet Distances}
\label{sec:gate-structure-details}
The channel histograms and hierarchy analysis use the L8 MLP gates of the SiT-XL/2 EMA checkpoints at 400K steps for Vanilla, Dispersive, and REPA (DINOv2-B), evaluated at $t=0.5$ for all 1,000 classes.

\paragraph{Gate histograms.}
We pool the signed gate entries over all $1{,}000\cdot1{,}152$ coordinates without centering, squaring, or normalization.
We use width-0.025 bins spanning $[-8,8]$, which cover every entry, and display $[-2,2]$.
Heights are percentages of all coordinates per bin, lightly smoothed with a Gaussian kernel of standard deviation one bin.

\paragraph{WordNet distances.}
For every class, we take the absolute value of the gate and center these magnitudes across the 1,000 classes separately for each channel.
We use WordNet 3.0 shortest hypernym-path lengths over all $\binom{1000}{2}=499{,}500$ unordered class pairs.
The Euclidean distance is computed between gate magnitudes, which class centering does not change, and the cosine distance between centered magnitudes.
We average distances over pairs within each WordNet distance bin, retaining bins with at least 500 pairs (distances 2--22).
The minimum bin mean of every curve occurs at WordNet distance 2. We divide each Euclidean curve by this minimum, which removes the difference in overall gate scale between models, and subtract it from each centered cosine curve, which is already scale-invariant.

\subsection{Gate Structure at Additional Timesteps}
\label{sec:gate-timestep-checks}
\begin{figure}[!htbp]
    \centering
    \plotfigure{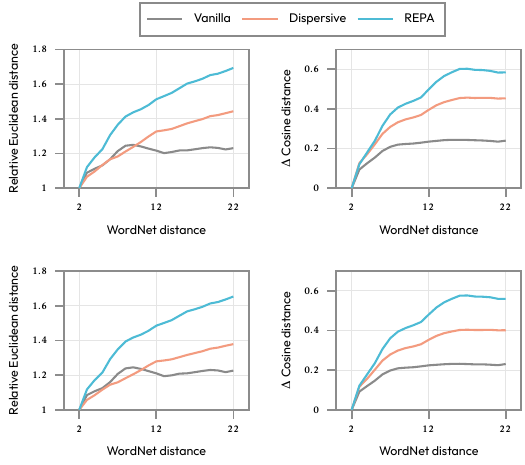}
    \caption{WordNet-distance analysis at additional timesteps, using the same checkpoints and distance definitions as \cref{fig:gate-wordnet-distance}. Top row: $t=0.25$; bottom row: $t=0.75$. Left: Euclidean distance of gate magnitudes. Right: cosine distance of class-centered gate magnitudes. Euclidean curves are divided by, and cosine curves shifted by, their own bin mean at WordNet distance 2, with the same retained WordNet bins as in the main text.}
    \label{fig:gate-wordnet-timesteps}
\end{figure}

\begin{figure}[!htbp]
    \centering
    \plotfigure{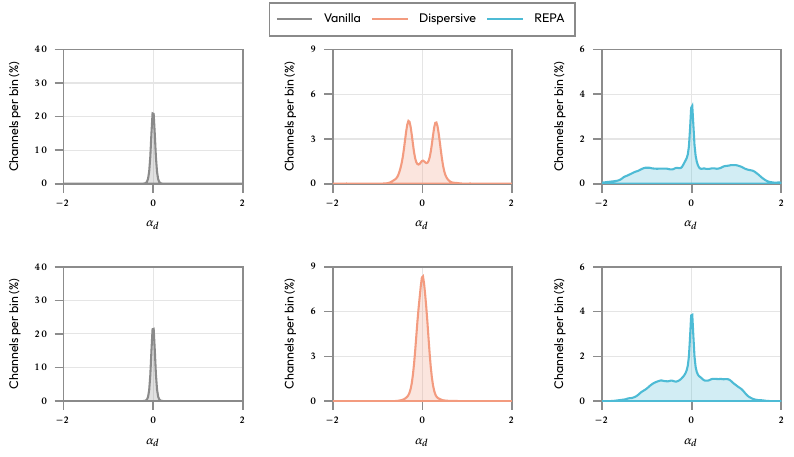}
    \caption{Signed gate histograms at additional timesteps, using the same checkpoints, bins, smoothing, and display range as \cref{fig:gate-channel-histogram}. Top row: $t=0.25$; bottom row: $t=0.75$. Columns: Vanilla, Dispersive, and REPA, with per-method y-axis limits shared across the two timesteps.}
    \label{fig:gate-histogram-timesteps}
\end{figure}

The WordNet-distance comparison and the gate histograms in the main text use $t=0.5$, the mean of the uniform timestep distribution used in training.
For the same checkpoints and L8 MLP gates, we additionally evaluate $t=0.25$ and $0.75$.
\cref{fig:gate-wordnet-timesteps} shows the same qualitative relationship at both timesteps: gate distances increase with WordNet distance, with REPA showing the largest separation across semantic distances.
\cref{fig:gate-histogram-timesteps} shows that the differences in gate scale between the three models persist at both timesteps.

\subsection{Teacher Sweep}
\label{sec:teacher-sweep-details}
We use eight REPA SiT-XL/2 EMA checkpoints at 400K steps released by \citet{SinghEtAl2026} on Hugging Face (\texttt{REPA-E/iREPA-collections}), with CLIP-L~\citep{RadfordEtAl2021}, DINOv2-B, DINOv3-B and DINOv3-H+~\citep{SimeoniEtAl2025}, LangPE-G and PE-G~\citep{BolyaEtAl2026}, MoCo-v3-B~\citep{ChenEtAl2021}, and WebSSL-1B~\citep{FanEtAl2025} teachers.
The FIDs are taken from the evaluations released in the same repository at step 400K, which use the ADM evaluator, CFG 1.0, and SDE sampling with 250 steps.
We measure \ourmetricmean for the L8 MLP gate of each checkpoint.

\subsection{Projector Sweep}
\label{sec:projector-sweep-details}
All projector variants use SiT-B/2 with DINOv2-B alignment after the fourth block (L4) and alignment coefficient 0.5.
Each variant is trained for 50K steps with the setup of \cref{sec:experimental-setting} and evaluated with the local FID protocol.

\paragraph{Projection heads.}
The three-layer MLP has widths $768\to1024\to1024\to768$, with SiLU between affine layers; the two-layer MLP has widths $768\to1024\to768$, with one SiLU.
The deep linear head uses the same widths as the three-layer MLP without intermediate nonlinearities, and the single linear head is $768\to768$.
Their parameter counts, including biases, are 2,624,256, 1,574,656, 2,624,256, and 590,592, respectively.
All variants normalize teacher targets within each image and channel as $(x-0.7\mu_N)/(\sigma_N+10^{-6})$, where $\mu_N$ and $\sigma_N$ are the mean and standard deviation over tokens.

\section{Details of the Radial Objective}
\label{sec:radial-details}

\subsection{Target Sweep and Target Choice}
\label{sec:radial-target-choice}
The radial penalty acts on the output of the final SiT-B/2 block (L12), flattened to a $256\cdot768$ dimensional vector per sample.
All sweep points use 100K-step EMA checkpoints trained with the setup of \cref{sec:experimental-setting} and evaluated with the local FID protocol.

\paragraph{Targets and coefficients.}
We search over $c\in\{875.885,1751.77,3503.54,4500,6000\}$.
For $N=256$ and $D=768$, $\sqrt{ND}\approx443.4$, and these targets correspond to $k\approx\{2,4,8,10,14\}$ in $c=k\sqrt{ND}$.
All five runs use $\lambda=6.032\times10^{-8}$, calibrated over the first five training iterations to make the weighted auxiliary gradient norm approximately 10\% of the velocity-loss gradient norm.
We select the target with the lowest FID.

\paragraph{Reference points.}
The Vanilla and Dispersive reference points are 100K EMA checkpoints evaluated with the same protocol; Dispersive uses coefficient 0.25 at the final block.

\begin{figure}[!htbp]
    \centering
    \plotfigure{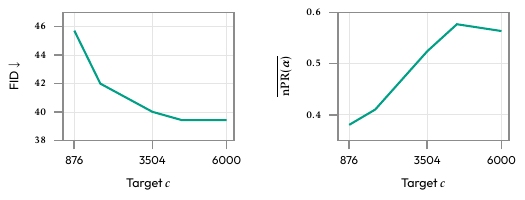}
    \caption{Sensitivity to the flattened feature-norm target for SiT-B/2 at 100K steps. Left: ADM FID. Right: \ourmetricmean for L12 MLP gates over 101 uniformly spaced model timesteps in $[0,1]$ and 1,000 classes.}
    \label{fig:radial-target-sensitivity}
\end{figure}

\paragraph{Sensitivity.}
\cref{fig:radial-target-sensitivity} plots FID and gate participation against the target.
The lowest target yields FID 45.72, close to Vanilla's 45.41; the target 3503.54 reaches 40.02, the selected target 4500 reaches 39.43, and the largest target 6000 reaches 39.44.
L12 MLP gate participation rises from 0.381 at the lowest target to 0.577 at the selected target and 0.564 at $c=6000$.

\subsection{Performance Comparison}
\label{sec:performance-protocol}
\cref{tab:performance-comparison} uses SiT-B/2 at $256\times256$, 400K training steps, the setup of \cref{sec:experimental-setting}, and the local FID protocol of \cref{sec:fid-protocol}.
The three teacher-free rows are the same runs as the training trajectories in \cref{fig:training-losses}.
Dispersive uses coefficient 0.25 at the final block, and the \ourmethod uses the target $c=10\sqrt{ND}\approx4434$ with coefficient $\lambda=5\times10^{-8}$.
REPA aligns L4 with DINOv2-B through a three-layer MLP projector with coefficient 0.5.
Mean gate participation is measured at L12 on the same EMA checkpoints.

\subsection{Training Dynamics}
\label{sec:training-dynamics-protocol}
\cref{fig:training-losses} uses the three 400K SiT-B/2 runs of \cref{sec:performance-protocol}.
The loss curves are smoothed with a trailing 5,000-step moving average over every logged iteration.
Each raw, unweighted auxiliary loss is normalized as $(L_t-L_{\min})/(L_0-L_{\min})$, with $L_{\min}=-\ln 32$ for Dispersive and $L_{\min}=0$ for Radial.
The gate participation panel evaluates the L12 MLP gate of the EMA checkpoints at 50K, 100K, 200K, and 400K steps with the protocol of \cref{sec:gate-measurement}.

%% file: sections/08-radial-analysis.tex
\section{Conditions and Proof of Proposition~\ref{prop:radial-channel-symmetry}}
\label{sec:radial-gate-analysis}

\paragraph{Analytical setting.}
Fix $(t,y)$ at the current model state and let $\vct{\alpha}=\vct{\alpha}_\theta(t,y)\in\mathbb R^D$.
We represent the input realizations at this condition by $B$ feature pairs $\mat{H}^{(b)},\mat{U}^{(b)}\in\mathbb R^{N\times D}$, held fixed with respect to the local gate variation, so that $\mathbb E[\,\cdot\mid t,y]$ in the main text is the average over $b$.
These realizations share $(t,y)$ and hence the gate; they need not appear together in a training minibatch.
Here $\radialnorm{b}$ denotes \cref{eq:radial-surrogate} evaluated at $(\mat{H}^{(b)},\mat{U}^{(b)})$.
The analyzed loss is
\begin{equation}
    \radialsurrogate(\vct{\alpha})=\frac1B\sum_{b=1}^{B}(\radialnorm{b}-c)^2.
    \label{eq:radial-surrogate-loss}
\end{equation}
Here $c>0$.
At the update point, all $\alpha_d$ are nonzero and all $\radialnorm{b}$ are positive.
These conditions ensure that the finite-average loss, \ourmetric, and the fractional gate updates are well-defined and smooth locally.
A positive auxiliary coefficient merely rescales the step size.

\paragraph{Precise channel-symmetry assumption.}
For every permutation $\pi$ of the $D$ channel indices, the two multisets
\begin{align}
    &\left\{\left(\|\mat{H}^{(b)}\|_{\mathrm F}^2,
    \|\mat{U}_{:,1}^{(b)}\|_2^2,\ldots,\|\mat{U}_{:,D}^{(b)}\|_2^2\right)\right\}_{b=1}^{B},\\
    &\left\{\left(\|\mat{H}^{(b)}\|_{\mathrm F}^2,
    \|\mat{U}_{:,\pi(1)}^{(b)}\|_2^2,\ldots,\|\mat{U}_{:,\pi(D)}^{(b)}\|_2^2\right)\right\}_{b=1}^{B}
\end{align}
coincide, including multiplicities.
The skip norm remains attached to its sample pattern, and the gate is not permuted.
This condition concerns squared-norm patterns, not individual feature entries or their signs.
It implies identical averaged squared channel norms, but is stronger than equality of these averages alone.
Neither Gaussianity nor independence is assumed.
Exact symmetry is a property of the analytical model, not a consequence of an architecture lacking hard-coded channel preferences and not a claim about observed mini-batches.

\paragraph{Relative gate update.}
Differentiating \cref{eq:radial-surrogate-loss} gives
\begin{equation}
    -\frac{\partial\radialsurrogate}{\partial\alpha_d}
    =\frac{2\alpha_d}{B}\sum_b\left(\frac{c}{\radialnorm{b}}-1\right)\|\mat{U}_{:,d}^{(b)}\|_2^2,
\end{equation}
which gives \cref{eq:radial-relative-update}.
Symmetry makes $B^{-1}\sum_b\|\mat{U}_{:,d}^{(b)}\|_2^2$ independent of $d$.

\paragraph{Ordering of the inverse-norm-weighted responses.}
For $i\ne j$, define the norm with the two channel norms interchanged:
\begin{equation}
\begin{split}
    \widetilde r_{\vct{\alpha}}^{(b;i\leftrightarrow j)}
    =\Big(&\|\mat{H}^{(b)}\|_{\mathrm F}^2
    +\alpha_i^2\|\mat{U}_{:,j}^{(b)}\|_2^2
    +\alpha_j^2\|\mat{U}_{:,i}^{(b)}\|_2^2\\
    &+\sum_{k\ne i,j}\alpha_k^2\|\mat{U}_{:,k}^{(b)}\|_2^2\Big)^{1/2}.
\end{split}
\label{eq:radial-swapped-norm}
\end{equation}
The channel-symmetry assumption gives
\begin{align}
    &\frac1B\sum_b\frac{\|\mat{U}_{:,i}^{(b)}\|_2^2}{\radialnorm{b}}
    -\frac1B\sum_b\frac{\|\mat{U}_{:,j}^{(b)}\|_2^2}{\radialnorm{b}}
    \nonumber\\
    &\quad=\frac1{2B}\sum_b
    (\|\mat{U}_{:,i}^{(b)}\|_2^2-\|\mat{U}_{:,j}^{(b)}\|_2^2)
    \left(\frac1{\radialnorm{b}}-\frac1{\widetilde r_{\vct{\alpha}}^{(b;i\leftrightarrow j)}}\right)
    \nonumber\\
    &\quad=-\frac{\alpha_i^2-\alpha_j^2}{2B}\sum_b
    \frac{(\|\mat{U}_{:,i}^{(b)}\|_2^2-\|\mat{U}_{:,j}^{(b)}\|_2^2)^2}
    {\radialnorm{b}\,\widetilde r_{\vct{\alpha}}^{(b;i\leftrightarrow j)}
    (\radialnorm{b}+\widetilde r_{\vct{\alpha}}^{(b;i\leftrightarrow j)})}.
    \label{eq:radial-swap-identity}
\end{align}
The summands in the final line are nonnegative.
Hence larger squared gate magnitudes have no larger inverse-norm-weighted responses, proving \cref{eq:radial-update-order}.
Equality of two squared gate magnitudes also implies equality of their weighted responses.
For unequal squared gate magnitudes, the ordering is strict if their squared channel norms differ in at least one sample.

\paragraph{One-step non-decrease of \ourmetric.}
For the proof only, define
\begin{equation}
    \vct{v}=-\nabla_{\vct{\alpha}}\radialsurrogate(\vct{\alpha}),\qquad
    \rho_d=\frac{v_d}{\alpha_d},\qquad
    p_d=\frac{\alpha_d^2}{\sum_j\alpha_j^2}.
\end{equation}
All $p_d>0$, and the prior nPR definition gives $\mathrm{nPR}(\vct{\alpha})=1/(D\sum_d p_d^2)$.
Differentiation in direction $\vct{v}$ yields
\begin{align}
    \left\langle\nabla_{\vct{\alpha}}\log\mathrm{nPR}(\vct{\alpha}),\vct{v}\right\rangle
    &=4\left(\sum_d p_d\rho_d-\frac{\sum_d p_d^2\rho_d}{\sum_d p_d^2}\right)
    \nonumber\\
    &=-\frac4{\sum_d p_d^2}\sum_{i<j}p_ip_j(p_i-p_j)(\rho_i-\rho_j)\geq0.
    \label{eq:radial-directional-npr}
\end{align}
Each summand has the stated sign by \cref{eq:radial-update-order}.
If the directional derivative is strictly positive, Taylor expansion gives
\begin{equation}
\begin{split}
    &\mathrm{nPR}(\vct{\alpha}+\eta\vct{v})-\mathrm{nPR}(\vct{\alpha})\\
    &\qquad=\eta\left\langle\nabla_{\vct{\alpha}}\mathrm{nPR}(\vct{\alpha}),\vct{v}\right\rangle+O(\eta^2)>0
\end{split}
\end{equation}
for every sufficiently small positive $\eta$.

It remains to justify finite-step non-decrease when the derivative is zero; a nonnegative first derivative alone would not suffice.
In that case, every summand in \cref{eq:radial-directional-npr} is zero.
Thus, $p_i\ne p_j$ implies $\rho_i=\rho_j$.
If $p_i=p_j$, the equal-gate case of \cref{eq:radial-swap-identity} also implies $\rho_i=\rho_j$.
Consequently all fractional rates have a common value $\rho$; hence $\vct{v}=\rho\vct{\alpha}$ and the step is exactly $\vct{\alpha}^{+}=(1+\eta\rho)\vct{\alpha}$.
This preserves \ourmetric whenever the scale is nonzero.
Together the two cases establish the finite-step inequality in Proposition~\ref{prop:radial-channel-symmetry}.

The increase is strict if there exist $i,j,b$ such that $\alpha_i^2\ne\alpha_j^2$ and $\|\mat{U}_{:,i}^{(b)}\|_2^2\ne\|\mat{U}_{:,j}^{(b)}\|_2^2$.
The argument concerns sufficiently small steps, not arbitrary learning rates.
Choosing the step small enough also preserves gate signs.

\paragraph{Quadratic comparison.}
For the same cross-term-excluded norm,
\begin{equation}
\begin{split}
    \mathcal Q_{\pm}(\vct{\alpha})
    &=\pm\frac1B\sum_b(\radialnorm{b})^2\\
    &=\pm\left[\frac1B\sum_b\|\mat{H}^{(b)}\|_{\mathrm F}^2
    +\sum_d\alpha_d^2\frac1B\sum_b\|\mat{U}_{:,d}^{(b)}\|_2^2\right].
\end{split}
\end{equation}
For this calculation only, let $\bar q$ denote the common average $B^{-1}\sum_b\|\mat{U}_{:,d}^{(b)}\|_2^2$.
A gradient step gives $\vct{\alpha}^{+}=(1\mp2\eta\bar q)\vct{\alpha}$ and therefore preserves \ourmetric whenever the scale factor is nonzero.\hfill\qedsymbol

\subsection{Scope of the Analysis}
\label{sec:radial-surrogate-scope}
The actual residual norm satisfies
\begin{equation}
\begin{split}
    &\|\mat{H}^{(b)}+\mat{U}^{(b)}\operatorname{diag}(\vct{\alpha})\|_{\mathrm F}^2\\
    &\qquad=(\radialnorm{b})^2
    +2\sum_d\alpha_d\langle\mat{H}_{:,d}^{(b)},\mat{U}_{:,d}^{(b)}\rangle.
\end{split}
\end{equation}
The analysis intentionally omits the last term.
Orthogonality such as $\vct{h}^{\top}\vct{u}=0$, or $\langle\mat{H},\mat{U}\rangle_{\mathrm F}=0$, does not imply that this gate-weighted cross term vanishes.
Nor does a small cross-term value at one gate imply a small derivative with respect to that gate.
The proposition applies to the cross-term-excluded norm, not to the full residual norm without additional assumptions.

In actual training the gate is parameterized; other features, task gradients, and optimizer state can also change.
The proposition concerns one local Euclidean gate update and does not imply monotonic \ourmetric growth under those full dynamics or a causal relationship between \ourmetric and FID.
The quadratic comparison is likewise restricted to the same analysis.
